\documentclass{styles/svproc}

\usepackage{comment} %
\usepackage{graphicx} %
\usepackage{bm} %
\usepackage{amsmath} %
\usepackage{amsfonts} %
\usepackage{enumitem}
\usepackage{booktabs}
\usepackage{siunitx}
\usepackage{booktabs}
\usepackage{subcaption}
\usepackage{hyperref}
\usepackage{tabularx}
\newcolumntype{Y}{>{\centering\arraybackslash}X}

\usepackage[style=ieee,citestyle=numeric-comp,natbib=true]{biblatex}
\usepackage{url}

\begin{document}
\mainmatter              %
\title{Learning Affordances from Interactive Exploration using an Object-level Map}
\titlerunning{Learning Affordances}  %
\author{Paula Wulkop\inst{1}$^{*}$ \and Halil Umut Özdemir\inst{1}$^{*}$
Antonia Hüfner\inst{1} \and Jen Jen Chung\inst{2} \and Roland Siegwart\inst{1} \and Lionel Ott\inst{1}}
\authorrunning{Paula Wulkop et al.} %
\tocauthor{Paula Wulkop, Halil Umut Özdemir, Antonia Hüfner, Jen Jen Chung, Roland Siegwart, and Lionel Ott}
\institute{Autonomous Systems Lab, ETH Zurich, Switzerland\\
\email{ pwulkop@ethz.ch, halilumutozdemir10@gmail.com}\\ 
$^{*}$ Equal contribution.\\
\and
School of EECS, The University of Queensland, Australia\\
This project has received funding from the European Union’s Horizon 2020 research and innovation programme under grant agreement No 101017008 (Harmony) and under the Marie Skłodowska-Curie grant agreement No 953454 (AERO-TRAIN).}

\maketitle              %

\begin{abstract}
Many robotic tasks in real-world environments require physical interactions with an object such as \textit{pick up} or \textit{push}. For successful interactions, the robot needs to know the object's affordances, which are defined as the potential actions the robot can perform with the object. In order to learn a robot-specific affordance predictor, we propose an interactive exploration pipeline which allows the robot to collect interaction experiences while exploring an unknown environment. We integrate an object-level map in the exploration pipeline such that the robot can identify different object instances and track objects across diverse viewpoints. This results in denser and more accurate affordance annotations compared to state-of-the-art methods, which do not incorporate a map. We show that our affordance exploration approach makes exploration more efficient and results in more accurate affordance prediction models compared to baseline methods.
\end{abstract}

\section{Introduction}
\label{sec:introduction}
Mobile robots operating in human spaces often need to interact with objects, for example for a task such as finding and retrieving a laptop, the robot needs to pick up the laptop before placing it on top of a table. To perform such manipulations of the environment, a robot must know the potential interactions it can perform with the objects in the world, which are known as affordances~\cite{gibson1977}. Most recent works on visual affordance estimation are supervised learning approaches that rely on pre-existing single-object affordance datasets ~\citep{myers2015affordance, do2018affordancenet, deng20213d, nguyen2016detecting, osama2022towards}. In contrast, we take the view that affordances for interactions are not only an object property but also depend on the robot's intrinsic capabilities~\cite{chung2022semantics}. For example, whether an object is pickupable depends on the strength and the gripper type of the robot arm. Therefore, a model learned using the experiences of one robot is not directly applicable to another robot with a different morphology. Additionally, to obtain good performance transfer between training and testing, the training data should ideally consist of objects contextualized in the environment and not of isolated objects, as is the case in most existing affordance datasets. Consequently, this work explores how to leverage self-supervised learning to create a robot-specific, scene-level affordance dataset without relying on manual labels.

\begin{figure}[t]
 \centering
    \includegraphics[width=.6\textwidth]{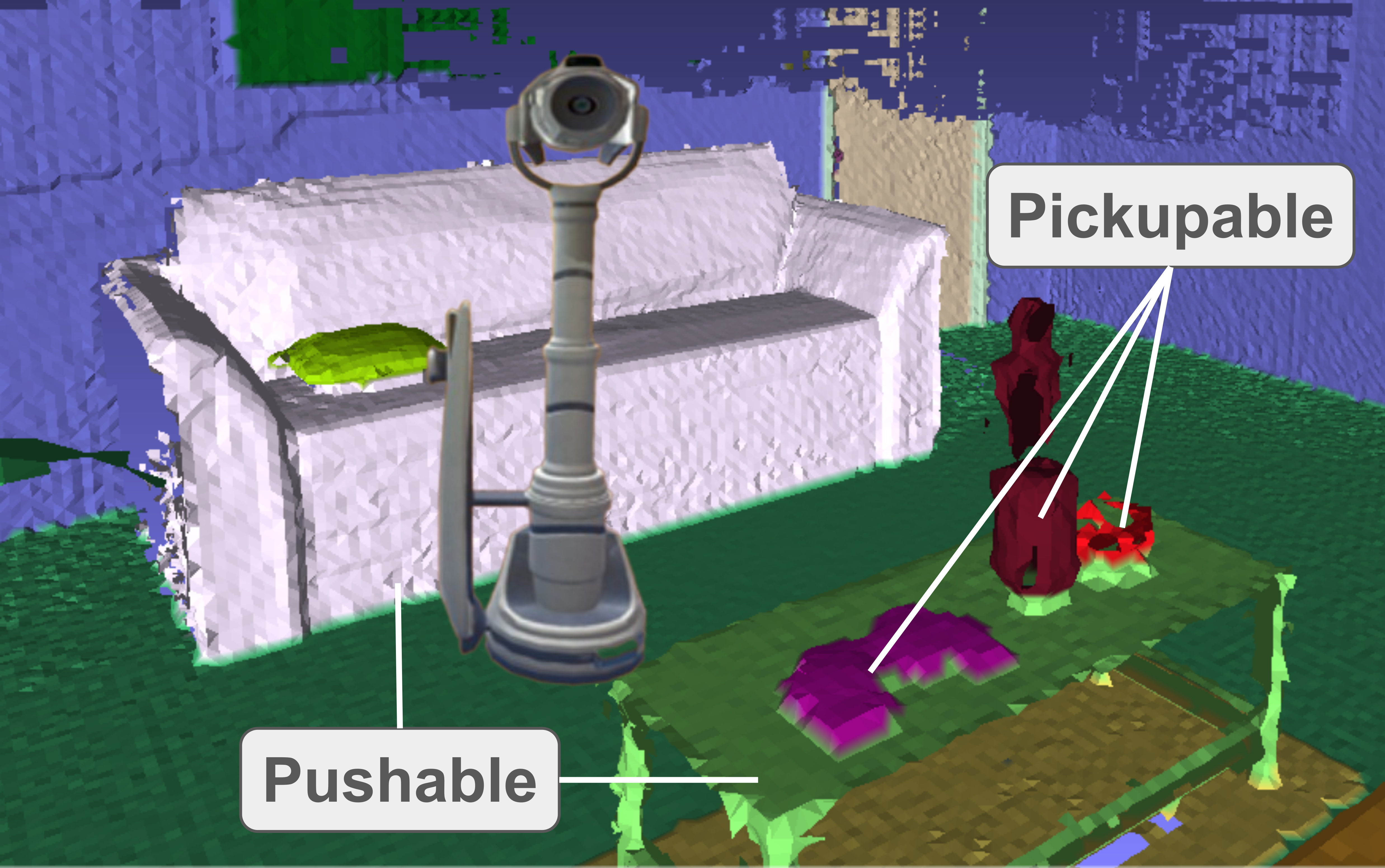}
\caption{An object-level map of a living room scene generated by TSDF++~\citep{grinvald2021tsdf++} with the robotic agent from the iTHOR simulation framework~\citep{kolve2017ai2thor}.}
\label{figure:ithor_env}
\vspace{-1mm}
\end{figure}

With this in mind, a promising approach is to learn affordances from data generated by a robot interacting with its environment in a photo-realistic simulator. Recently, \citet{nagarajan2020learning} proposed such an approach for learning affordances from interactions. They use deep reinforcement learning (RL) to explore a new 3D environment while, in parallel, training an affordance model based on the results of the executed interactions. The authors showed that affordance-based exploration leads to faster discovery of interactable regions. Additionally, their learned affordance model proved helpful in complex downstream tasks like washing the dishes. However, since the exploration framework of~\citet{nagarajan2020learning} operates on each frame individually, it is unable to exploit structural scene information to expedite the agent's affordance exploration and learning, i.e. the learned exploration agent has no sense of object permanence. In contrast, modern robotic mapping pipelines provide representations at the object level~\citep{grinvald2021tsdf++, schmid2022panoptic} and explicitly offer dense object segmentations that can be associated across multiple viewpoints. This information can be directly applied to extrapolate observed interaction data across the full exploration sequence and substantially improve learning efficiency by providing higher interaction success rates.

In this work, we create and use an explicit map of the environment in which each discovered object and interaction is stored (see Fig.~\ref{figure:ithor_env}). We integrate TSDF++~\citep{grinvald2021tsdf++}, a dynamic object-level mapping framework, into the pipeline to enable the agent to identify different object instances and track their movements across different viewpoints. Our approach consists of an RL agent exploring the simulated training environments by interacting with various objects and storing this data in the object-level map. From this, we periodically extract a dataset to concurrently train an affordance classifier that learns to accurately predict agent-specific object affordances. While the RL policy ensures that the robot efficiently explores the scene during training, the final output of the system is the learned affordance predictor.

Our contribution is the introduction of an object-level map into an interactive affordance learning pipeline. The benefits of an explicit map representation are twofold: Firstly, storing the interaction experiences in the object-level map permits the propagation of interaction outcomes across different viewpoints, creating more accurate and complete annotations. Secondly, explicitly including object information as state information improves the interaction success rate and efficiency, leading to higher-quality training data. Overall, we show that the usage of an object-level map improves the quality of the affordance predictions, therefore allowing a robot to explore a new scene more efficiently.

\section{Related work}
\label{sec:related work}

\subsection{Visual Affordance Learning} 
Affordances are defined as the actions that an agent can apply to an object \citep{jamone2016affordances, gibson1977}. They are usually learned from visual inputs by either segmenting the objects and extracting the parts that are relevant for an interaction \citep{kokic2017affordance, li2023oneshotopenaffordancelearning}, or by performing dense, pixel-wise affordance predictions \citep{myers2015affordance, do2018affordancenet, deng20213d, nguyen2016detecting}. Most state-of-the-art approaches focus on affordance learning via supervised learning, for which annotated datasets of object affordances are required~\citep{myers2015affordance, do2018affordancenet, deng20213d, nguyen2016detecting}. However, most datasets only contain single-object affordances \citep{koppula2013learning, myers2015affordance, nguyen2017object, osama2022towards} and do not consider the context in a scene. Additionally, due to the physical constraints of each robot, affordance definitions can vary from robot to robot. Therefore labels extracted either from human annotation or from demonstrations~\citep{zhu2016inferring, nagarajan2019grounded, fang2018demo2vec, alayrac2017joint, nagarajan2020egotopo, 2024scenefun3d} do not always generalize across different robots. Rather than learning affordances from annotated datasets, we therefore propose to learn affordances from data generated by an agent interacting with different objects in a scene-level simulation environment.

\subsection{Interactive Exploration} 
Most works on the exploration of unknown environments focus on navigation actions, using either learning-based methods like reinforcement learning \citep{nguyen2019reinforcement, tan2019learning, chen2019learning, ramakrishnan2021exploration, qi2020learning} or classical SLAM-based methods \citep{talbot2020robot, chaplot2020learning}. Recently, \citet{liang2023alp} proposed an embodied learning framework that uses the data collected during active exploration of a 3D environment to learn visual representations. However, not only navigation but also interactions with objects, e.g. opening the fridge or picking up a pillow, are important for using robots in our daily lives. Recently developed simulators with high-level interaction capabilities~\citep{kolve2017ai2thor, puig2018virtualhome, gao2019vrkitchen} enable pipelines where agents can follow instructions or answer questions that require interaction~\citep{shridhar2020alfred, gao2019vrkitchen, das2018embodied, gordon2018iqa}. While these works focus on task-driven goals, we aim to use the interactions in order to build up an explicit knowledge of affordances.

\subsection{Map Representations for Exploration}  
A structured representation of visited locations is beneficial for effectively exploring a new environment. \citet{gervet2022navigating} show that the usage of explicit maps as a state compared to an implicit memory improves the generalization performance of the agent. \citep{ramakrishnan2021exploration, chen2019learning} use a top-down occupancy map to keep track of explored areas and the obstacles in them. In \citep{zhang20233daware} 3D information is crucial since the goal is to find a specific object in a scene, so they use fused 3D point clouds as observations. Similarly to our approach, \citet{chaplot2021seal} build a 3D semantic map for self-supervised label propagation. Specifically, the labels are stored in a 3D semantic map and then projected onto the agent’s frame-wise observations from different viewpoints, thereby generating pixel-wise instance labels for new frames. In this work, we will use a similar approach to generate additional affordance labels.

\subsection{Learning Affordances from Interactions} 
In previous works, interactive exploration has been used to learn affordances for simple objects in table-top environments \citep{legoff2022, borjadiaz2022} and for block pushing tasks \citep{kim2015}. \citet{Wang_2024_CVPR} focus on learning affordances for moving obstacles out of the way when navigating through a cluttered scene. Most relevant to our work, \citet{nagarajan2020learning} recently proposed a method to interactively learn generic object affordances on a scene-level. Their approach consists of a reinforcement learning algorithm that learns the interactive exploration strategy and an affordance network that is trained with the interactions executed during exploration. For each executed interaction, they mark the point of interaction and back-project it to the previous frames in order to annotate objects from different viewpoints. However, since they do not create an explicit map, they cannot re-identify an object instance, leading to very sparse annotations and a low interaction success rate. 

\section{Approach}
\label{sec:approach}

\begin{figure}[ht]
    \centering
    \includegraphics[width=.95\textwidth]{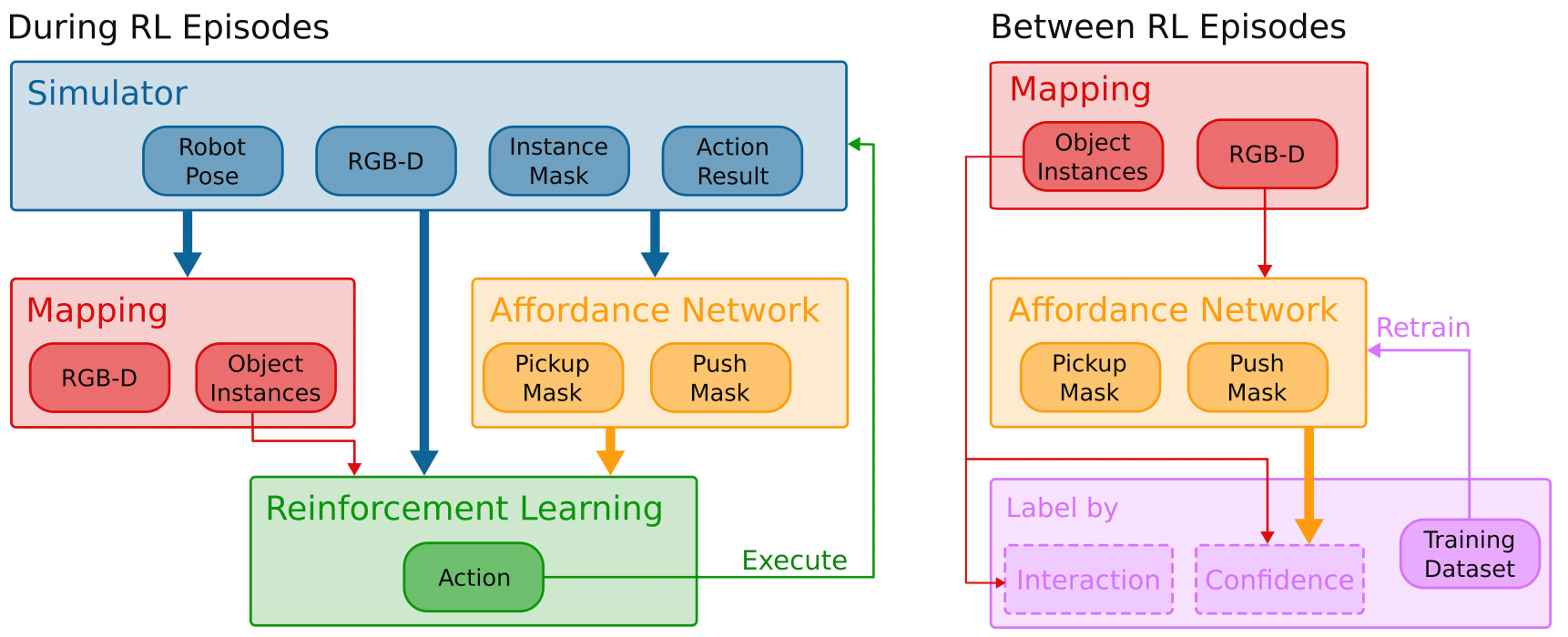}
    \caption{\textbf{Overview of our method during training.} At each time step, an action is executed and the simulator (blue) outputs if the action was successful, as well as the RGB-D image, ground truth instance segmentation mask, and robot pose. The mapping module (red) updates the map with this data, while the affordance module (yellow) predicts the affordances. The RL exploration policy module (green) estimates the next optimal action using the current state of the object-level map, RGB-D image, and affordance estimation as input. At the end of the episode, the label module (purple) annotates each object instance based on the interaction data and estimations from the current affordance network. Finally, the affordance network is retrained after every episode with the new data.}
    \label{figure:methodology_summary}
     \vspace{-1mm}
\end{figure}

Our approach aims to train an affordance model using an agent's previous experiences, which are stored in an object-level map. Fig.~\ref{figure:methodology_summary} shows an overview of our approach. The first component on the top shows the simulation environment where the RL agent can collect experiences by moving around and attempting to either \textit{pick up} or \textit{push} the objects in the scene. The simulator then returns the result of whether the interaction was successful or not. Once an RL episode finishes, the collected experience, in conjunction with the object level map and the RGB-D data from the onboard sensor, is used to generate new labels to improve the affordance network. In the following, we provide detailed descriptions of these components.

\subsection{Mapping Framework}
An object-level map is used to reconstruct the environment during exploration. At every step, the RGB-D frame and the instance segmentation mask are used to update the map. Additionally, the affordance label is stored in the map for those objects which have already been annotated. We use TSDF++~\citep{grinvald2021tsdf++} which is a voxel-based mapping framework that models each object instance as a separate truncated signed distance function (TSDF) layer. An important feature of TSDF++ is that it is a dynamic object-level framework, allowing it to track object movements when the agent interacts with an object. The advantage of tracking the object is that the affordance label can be applied to the same object observed from different viewpoints. One limitation of TSDF++ is that it can only be used with rigid objects, i.e. affordances such as \textit{slice} or \textit{open} cannot be represented in the map. With a more flexible mapping framework however, the approach presented here would also extend to such actions.

\subsection{Interactive Exploration Policy}\label{sec:interactive_exploration_policy}
We model this exploration task as a Markov decision process and use reinforcement learning to learn the best policy. The action space, state space, and reward are described in the following.

\begin{figure}[t]
    \centering
    \includegraphics[width=.6\textwidth]{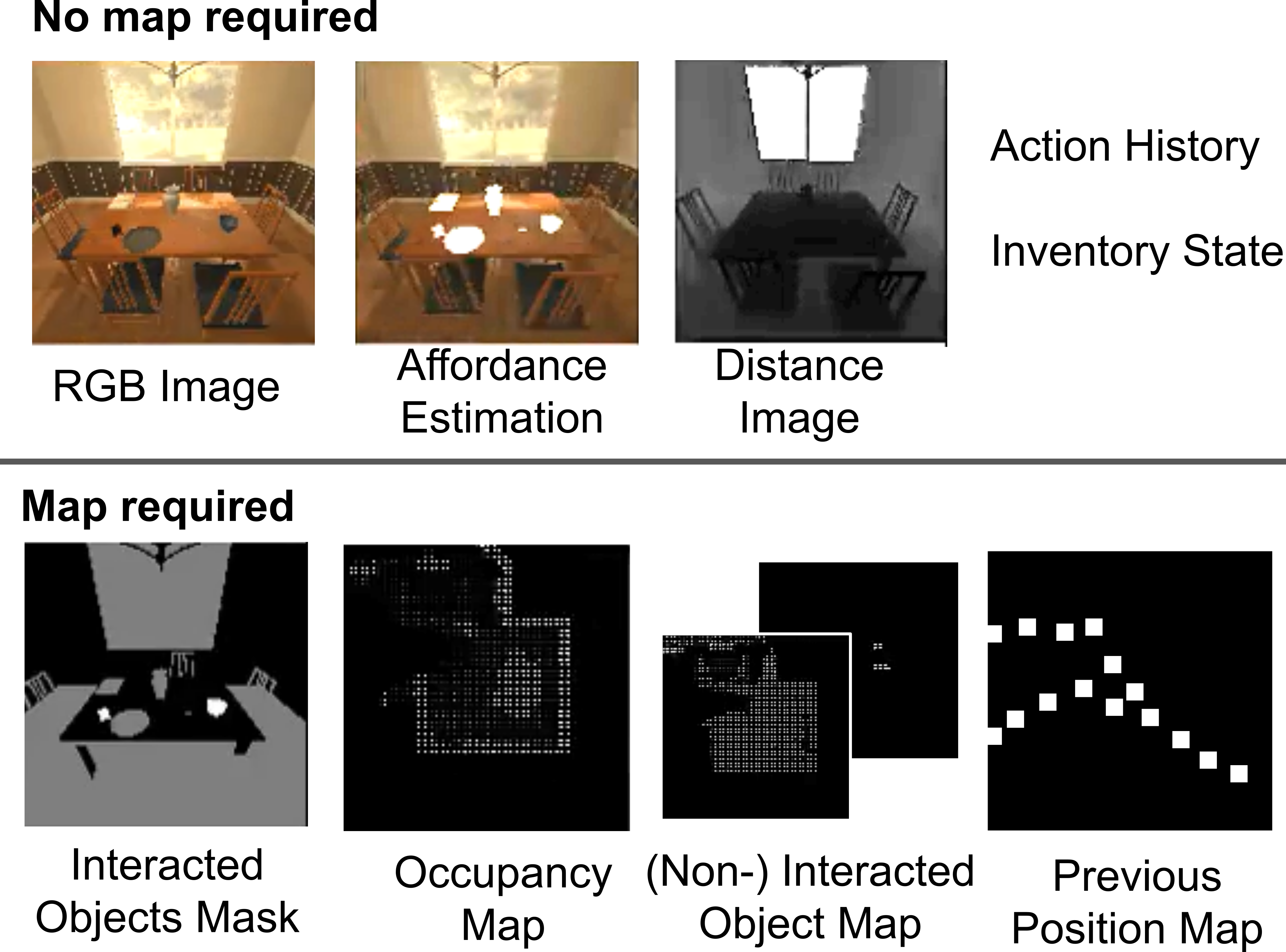}
    \caption{\textbf{Visual representation of the state space.} The elements in the top row do not require a map, while the states in the bottom are obtained through the map.}
    \label{figure:state_space}
     \vspace{-1mm}
\end{figure}

\paragraph*{State space}
The state space of the agent is a combination of the raw perception input, affordance estimation, spatial representations, and action history. Fig.~\ref{figure:state_space} shows the visual representation of the state space. From the onboard sensor the RGB image as well as the depth image are used, from which we create a so-called distance image by projecting the depth image into the robot arm's base frame to indicate if an object is within reach. The output of the current affordance estimation network indicates whether an object is likely to be interactable. Additionally, the state space contains the action history of the last five actions with their obtained success and reward as well as the inventory indicating if the robot currently holds a picked up object. In addition to these states derived from the current RGB-D input, we add states based on the object-level map to improve exploration efficiency. An egocentric 2D grid map of the previously visited positions as well as the interacted objects mask with the labels \textit{interaction successful}, \textit{interaction unsuccessful} and \textit{no interaction attempted} are used as states. Furthermore, we create an occupancy map and an interacted and a non-interacted objects map by projecting the 3D voxels into 2D. While the compression into 2D space, which is necessary to ensure computational feasibility of the RL pipeline, inevitably leads to a loss of information about the vertical arrangement of objects, it captures the planar layout which is crucial for navigation. All maps are used once as global maps with a size of $\SI{10}{\meter} \times \SI{10}{\meter}$ and once as local egocentric maps of size $\SI{2}{\meter} \times \SI{2}{\meter}$.

\paragraph*{Action space}
The action space of the robot in the iTHOR \citep{kolve2017ai2thor} simulation environment consists of navigation actions and interactions with objects. The discrete navigation actions are: \emph{forward movement by \SI{0.25}{\meter}}, \emph{rotate right by $\SI{30}{\degree}$}, \emph{rotate left by $\SI{30}{\degree}$}, \emph{look up by \SI{15}{\degree}}, \emph{look down by \SI{15}{\degree}}. The possible interactions with objects are \textit{pick up}, \textit{drop} and \textit{push}. Once the robot picks up an object, it can attempt to move the object by $\pm \SI{0.1}{\meter}$ along the three axes.

\paragraph*{Interaction selection}
For every robot pose there are usually multiple objects that the robot could try to interact with. \citep{nagarajan2020learning} solve this ambiguity by always selecting the object that is most centered in the current RGB frame. We however want to enable the robot to actively select the object to interact with as we believe that this will lead to a more targeted exploration. Since the number of objects varies from frame to frame but the RL action space has a fixed size, we implemented the following approximation for the object selection: The center region of the image frame is split up in a grid of nine equal-sized cells and the objects are assigned to the grid cell that they are closest to. In this way, the robot selects the object by selecting the corresponding grid cell, where empty grid cells cannot be selected due to a mask and for cells with multiple objects the largest object will be selected. Additionally, we use the output of the affordance network to filter out objects with a low confidence score. 

\paragraph*{Reward}
To train the agent, we use a reward function with three components, as shown in \eqref{eq:reward}. $R_\text{nav}$ encourages spatial exploration by rewarding each newly visited position as well as each new orientation at a previously visited position. $R_\text{int}$ rewards each interaction with a new object instance, preventing exploitation of known objects. $R_\text{fail}$ punishes each failed action, for example if the agent tries to pick up a couch which is non-pickupable, discouraging repeat negative interactions. 
\vspace{-6pt}
\begin{align}
\label{eq:reward}
    R & = \alpha_1 R_{\text{nav}} + 
        \alpha_2 R_{\text{int}} +
        \alpha_3 R_{\text{fail}} \\
    R_{\text{nav}} & = \begin{cases}
        1 & \text{if novel position} \\[-3pt]
        0.3 & \text{if novel orientation at prev. pose} \\[-3pt]
        0 & \text{otherwise}
    \end{cases} \\
    R_{\text{int}} & = \begin{cases} 
     1 & \text{if successful new interaction} \\[-3pt]
        0 & \text{otherwise}
    \end{cases} \\
    R_{\text{fail}} & = \begin{cases} 
         -1 & \text{if action failed} \\[-3pt]
        0 & \text{otherwise}
        \end{cases}
\end{align}
\vspace{-6pt}

\paragraph*{Exploration policy learning}
The policy network is trained using Proximal Policy Optimization (PPO) \citep{schulman2017proximal} with rollouts of 600 time steps.
All the elements of the state space are encoded using 2D convolutional encoders for the image-like inputs with a dimension of (128, 128) and MLPs for 1D states. These encodings are concatenated and then passed through two three-layer MLPs to separately predict the action function and the value function, with MLP layer dimensions of (64, 32, 21) and (64, 32, 1), respectively. Additionally, to obtain faster convergence, we apply an action mask which disables actions that are infeasible given the current state, e.g. the action \emph{drop} is disabled if the agent does not hold any object.

\subsection{Affordance Learning}
The affordance estimation network is trained on the interaction data collected during exploration, as shown in Fig.~\ref{figure:methodology_summary}. In the following, we will explain how the interaction data collected by the robot during an episode is labeled and used to retrain the affordance model. The improved affordance model is then used in the exploration pipeline, thereby ensuring that the two components are helping each other to converge.

\begin{figure}[t]
  \centering
  \includegraphics[width=0.7\textwidth]{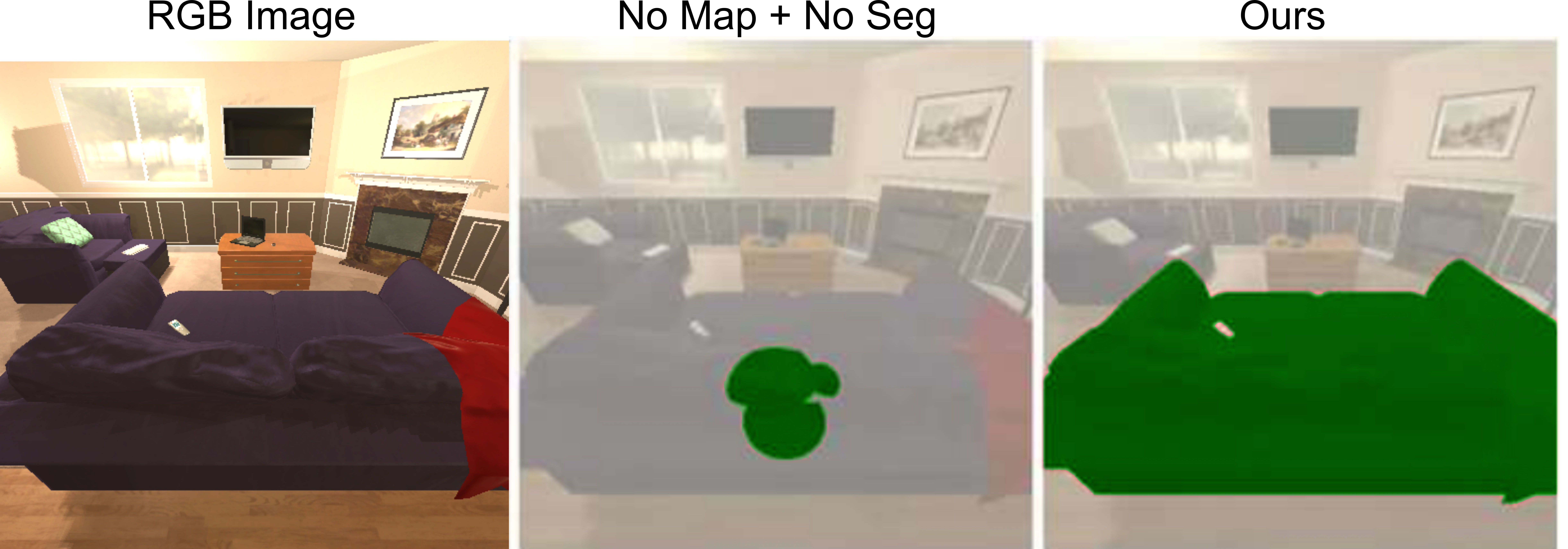}
  \caption{An example image of the annotation approach used by \citep{nagarajan2020learning} and the No Map + No Seg ablation (middle) compared to our approach which uses object segmentation and is therefore able to annotate the full object (right). Annotation masks are shown in green.}
  \label{figure:annotation_strategy}
\end{figure}

\paragraph*{Data generation and labeling}
To generate data, the exploration policy executes at each RL time step an action and the simulator returns whether it was successful or not. Rather than annotating the object directly on a single RGB-D frame after each interaction, we annotate the object instance in the map. This method is similar to SEAL~\citep{chaplot2021seal} which also uses a map to propagate annotations. The labelled map allows us to annotate each frame of the episode in which this object instance has been visible, thereby applying the information of a single interaction to multiple frames. For each frame we use the segmentation mask to create dense labels for all pixels of an annotated object. In our implementation, the segmentation mask comes directly from the simulator, but it could also easily be obtained by a separate state-of-the-art segmentation network, e.g.~\citep{SAM}. The baseline \citep{nagarajan2020learning} instead does not use a segmentation mask and instead labels a sphere of radius \SI{20}{\centi\meter} centered on the interaction position (see Fig.~\ref{figure:annotation_strategy}). Since this sphere is of a fixed size, this strategy can lead to false positive labels for small objects and false negative labels for large objects.

Overall, the instance mapping allows us to consistently annotate objects across different viewpoints and robot movements and thereby creating more densely labeled datasets. The first and most straightforward annotation approach is to perform \textit{annotation by interaction}, meaning that if the agent performed a successful or unsuccessful interaction with an object instance, it is annotated accordingly. However, since these interactions are sparse, we propose to additionally apply a self-supervised annotation strategy which we call \textit{annotation by confidence}. This approach takes advantage of the classifier likelihoods outputted by the affordance network to label additional data. After the affordance network starts to converge, the affordance prediction is queried for each frame $F_i$ and is used to further annotate objects with which the robot has not interacted. For each frame $F_i$ in which the object appears, the percentiles of the affordance predictions are computed. If the maximum of the $95$th percentile over all these frames is higher than a threshold $\xi_T = 0.9$, or if the minimum of the $5$th percentile is lower than $\xi_F = 0.1$, then the object instance is annotated accordingly as a successful or unsuccessful interaction, i.e.:
\vspace{-3pt}
\begin{equation}
    \text{label} = \begin{cases}
        1 & \text{if} \max\left\{
            \left\{ F_i \right\}_{P95} \forall i
        \right\} > \xi_T \\[-3pt]
        0 & \text{if} \min\left\{
            \left\{ F_i \right\}_{P5} \forall i 
        \right\} < \xi_F \\[-3pt]
        \text{none} & \text{otherwise.}
    \end{cases}
\end{equation}
Note that since neural networks are known to
be over-confident in their predictions, it could be beneficial to apply a calibration method, as suggested in \citep{guo2017}. At the end of each episode a partially annotated RGB-D affordance dataset is created by combining the object interaction annotations, RGB-D images and segmentation masks for each frame. In order to ensure a balanced dataset, we only use frames that contain pixels with both positive and negative labels.

\paragraph*{Training procedure}
The affordance network is based on a U-Net architecture~\citep{ronneberger2015unet} with a loss function that is a combination of binary cross entropy and dice loss. The U-Net implementation contains an encoder and a decoder with 4 downscale layers each, with output sizes of (24, 48, 92, 164). To train the affordance network, we use the Adam optimizer with weight decay. After each episode, the newly generated dataset is split up into training, validation, and testing sets and added to a global affordance dataset. The training of the affordance network is then continued with this global dataset, whereby the dataset of an episode is removed from the global dataset after its $35$th usage in the training loop. If the newly trained affordance network outperforms the previous one, it replaces it.

\section{Experiments}
\label{sec:experiments}
In this section we evaluate the effect of using an object-level map for the affordance learning. First, we evaluate if it increases the efficiency of the policy learning. Second, we evaluate the accuracy of the resulting affordance model. 

\subsection{Experimental Setup}

\paragraph*{Simulation environment}
We evaluate our pipeline on the iTHOR~\citep{kolve2017ai2thor} simulation environment, which supports high-level interactions with various types of objects in realistic 3D indoor scenes. We use living room scenes such as the one shown in Fig.~\ref{figure:ithor_env} since they contain a large variety of objects that the agent can interact with and also have large areas to explore. The scenes contain up to 50 different object categories which support 0 to 2 interactions (\textit{push}, \textit{pick up}). We split the 30 living room scenes into 25 training scenes (20\% of this data is used for validation) and five testing scenes.

\paragraph*{Baseline}
 We compare our method to the best-performing method in the state-of-the-art paper by \citet{nagarajan2020learning} which they call ``IntExp(PT)". Note that they also evaluated a different version of their method that uses object segmentation instead of spherical labels (``IntExp(Obj)"), but this method was outperformed by ``IntExp(PT)". For this reason, we compare against the ``IntExp(PT)" approach, which we implemented in our framework and call it here \-IntExp~\citep{nagarajan2020learning}. The following adaptions were necessary to make their approach directly comparable to our results: We retrained their code on the two affordances \textit{pick up} and \textit{push}, since they had used additional affordances which are not covered by our approach. Additionally, we used the AI2-THOR living room environments instead of the kitchen and we increased the image size slightly from 80x80 pixels to 128x128 pixels, since these settings were also used in our approach. Finally, our computational hardware - which we also used to train our approach - limited us to run 8 parallel processes during the RL training instead of 16 as in the original implementation, while still keeping the total number of training steps the same.

\paragraph*{Ablations}
To evaluate how the integration of an object-level map impacts the affordance prediction and the exploration rate, we compare our approach to two ablations of our full method:
\vspace{-6pt}
\begin{itemize}[leftmargin=10pt]
\setlength{\parskip}{0pt}
\setlength{\itemsep}{0pt plus 1pt}
\item Ours: The full method using the object-level map as described in Section~\ref{sec:approach}. It uses the ground truth segmentation mask for labeling and applies annotation by interaction and confidence.  

\item No Map + Seg: This ablation evaluates the contribution of the map, therefore the state space contains only elements that do not require a map (\text{RGB image}, \text{distance image}, \text{affordance estimates}). As in IntExp~\citep{nagarajan2020learning}, instead of a map we use a Gated Recurrent Unit (GRU) module in the RL part to aggregate observations over time. 

\item No Map + No Seg: This ablation aims to evaluate the importance of object segmentation. Instead of the ground truth segmentation mask used in our full approach, we apply the annotation strategy from IntExp~\citep{nagarajan2020learning}. Annotations from multiple viewpoints are generated by projecting 3D interaction points obtained during the episode (successful and unsuccessful) into all captured images where the object was visible shortly before and after the interaction. As objects have a volume, a sphere of radius \SI{20}{\centi\meter} centered on these interactions is used to generate interaction labels. 

\end{itemize}

\subsection{Affordance-based Interactive Exploration}
In this section, we address the question of whether the interactive exploration strategy is made more efficient by using an explicit map. To evaluate the performance during training of our pipeline compared to the ablation baselines without a map, we use the following metrics:

\vspace{-6pt}
\begin{itemize}[leftmargin=10pt]
  \setlength{\parskip}{0pt}
  \setlength{\itemsep}{0pt plus 1pt}

\item Affordance IoU: Intersection over Union of affordance predictions compared to ground truth labels per frame.
\item Object-level Accuracy: The ratio of objects with correctly predicted affordance labels over all visible objects in a frame. Object-level labels are obtain by performing a majority vote of all pixels within the object boundaries.
\item Interaction Success Rate: The ratio of successful interactions over all attempted interactions during an episode.
\item Interacted Object Rate: The ratio of the objects that the robot interacted with during an episode over all interactable objects in the scene.
\item Interactable Annotation Rate: The ratio of pixels that are correctly annotated as \textit{interactable} over all pixels.
\item Non-interactable Annotation Rate: The ratio of pixels that are correctly annotated as \textit{non-interactable} over all pixels.
\end{itemize}
\vspace{-6pt}

\begin{figure}[t]
  \centering
   \begin{subfigure}[b]{0.32\textwidth}
        \centering
         \includegraphics[width=\textwidth]{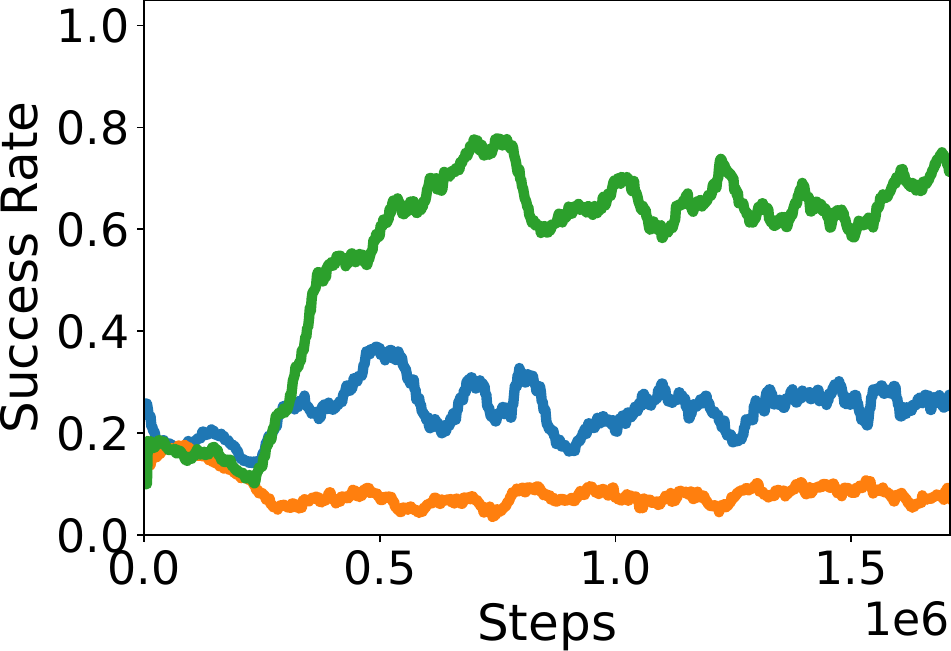}
        \vspace{-6mm}
         \caption{Interaction Success Rate}
         \label{fig:int_succ}
         \end{subfigure}
        \hfill
    \begin{subfigure}[b]{0.32\textwidth}
         \centering
         \includegraphics[width=\textwidth]{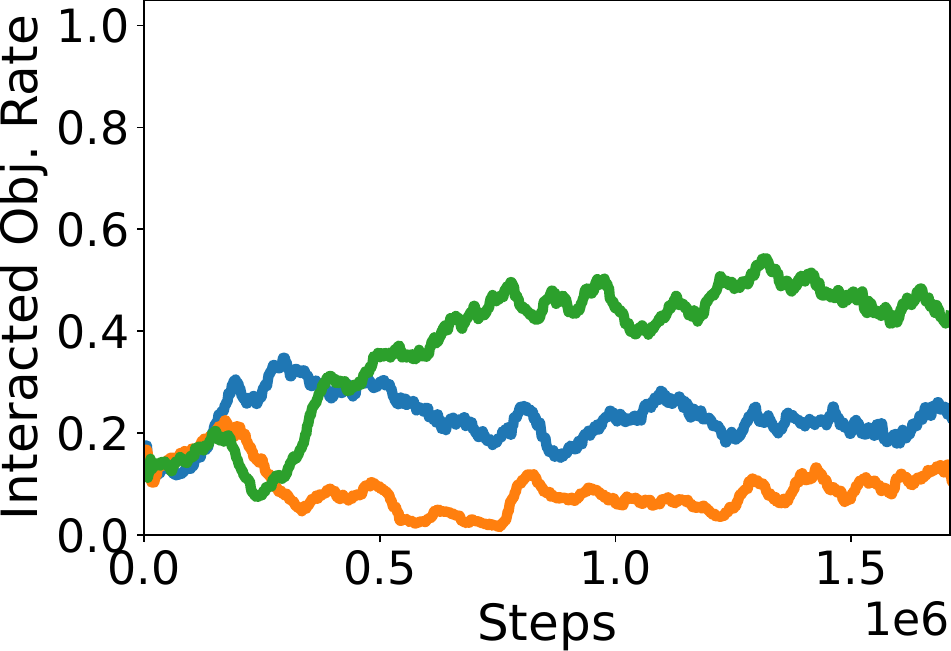}
        \vspace{-6mm}
         \caption{Interacted Object Rate}
        \label{fig:int_obj}
         \end{subfigure}
        \hfill
    \begin{subfigure}[b]{0.32\textwidth}
         \centering
         \includegraphics[width=\textwidth]{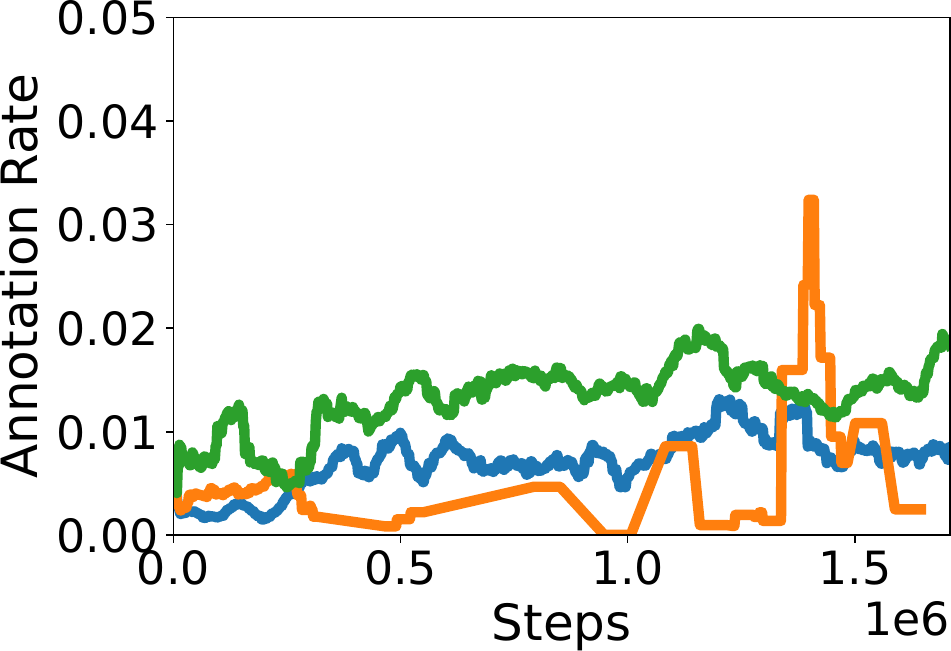}
        \vspace{-6mm}
         \caption{Interactable Annotat. Rate}
         \label{fig:annotation_rate_pos}
         \end{subfigure}
        \hfill
    \begin{subfigure}[b]{0.32\textwidth}
         \centering
         \includegraphics[width=\textwidth]{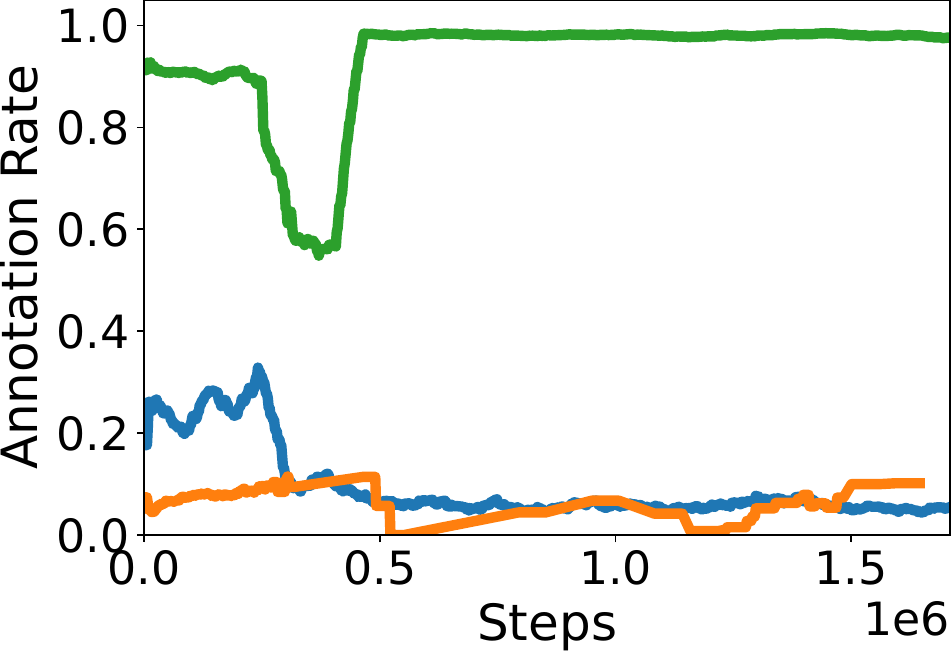}
        \vspace{-6mm}
         \caption{Non-interact. Annotat. Rate}
        \label{fig:annotation_rate_neg}
         \end{subfigure}
        \hfill
  \begin{subfigure}[b]{0.32\textwidth}
         \centering
         \includegraphics[width=\textwidth]{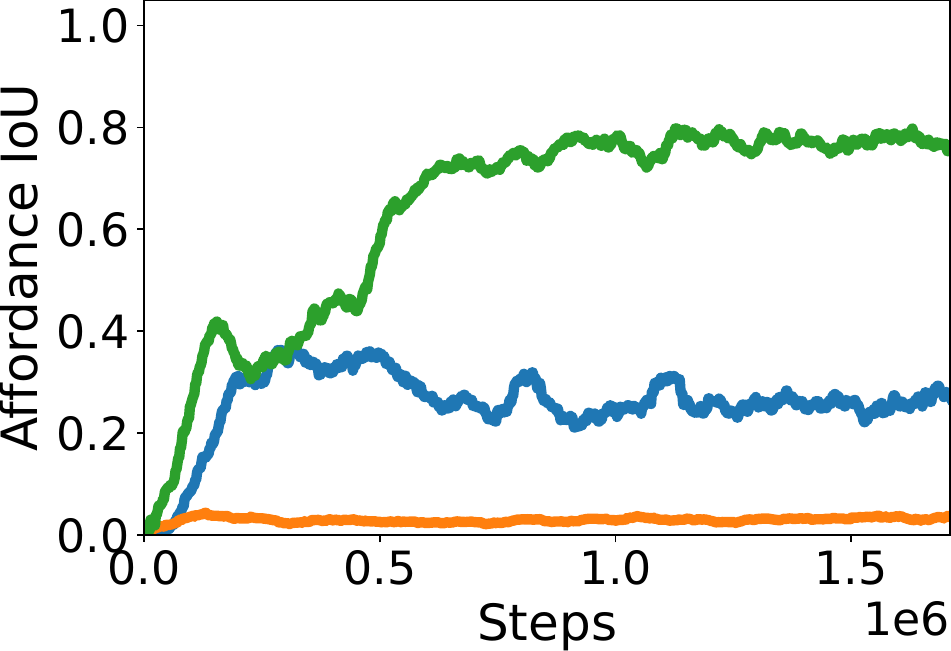}
         \vspace{-6mm}
         \caption{Affordance IoU}
        \label{fig:iou}
         \end{subfigure}
          \hfill
    \begin{subfigure}[b]{0.32\textwidth}
         \centering
         \includegraphics[width=\textwidth]{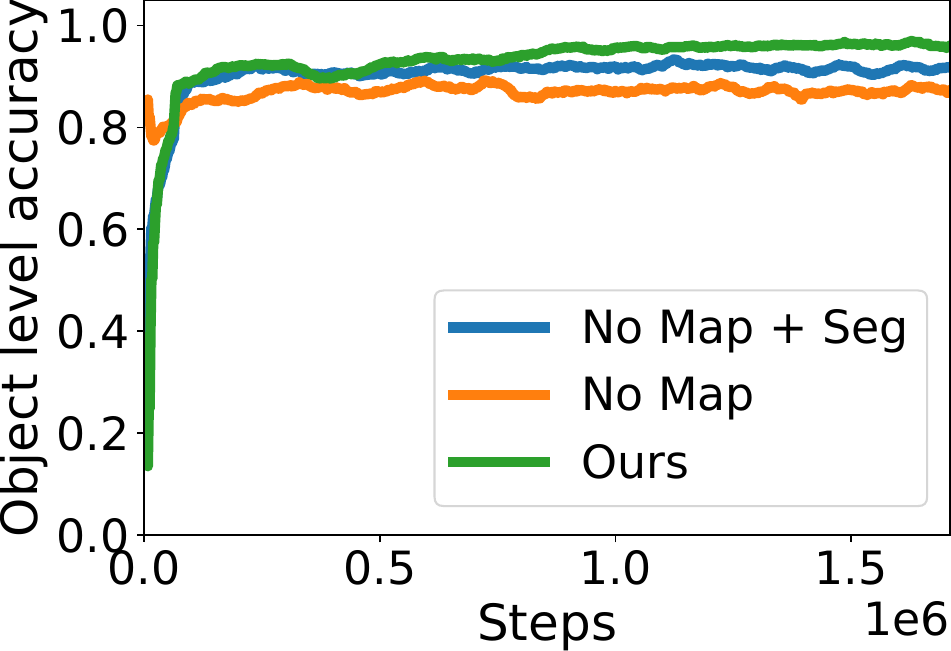}
        \vspace{-6mm}
         \caption{Object-level Accuracy}
         \label{fig:OAcc}
        \end{subfigure}
        \hfill
\caption{The training curves for the \textit{pick up} affordance show that our approach leads to a higher interaction success rate and a better affordance estimation performance.}
\label{figure:trainingcurves}   
\end{figure}

\begin{figure}[ht]
    \centering
    \includegraphics[width=\textwidth]{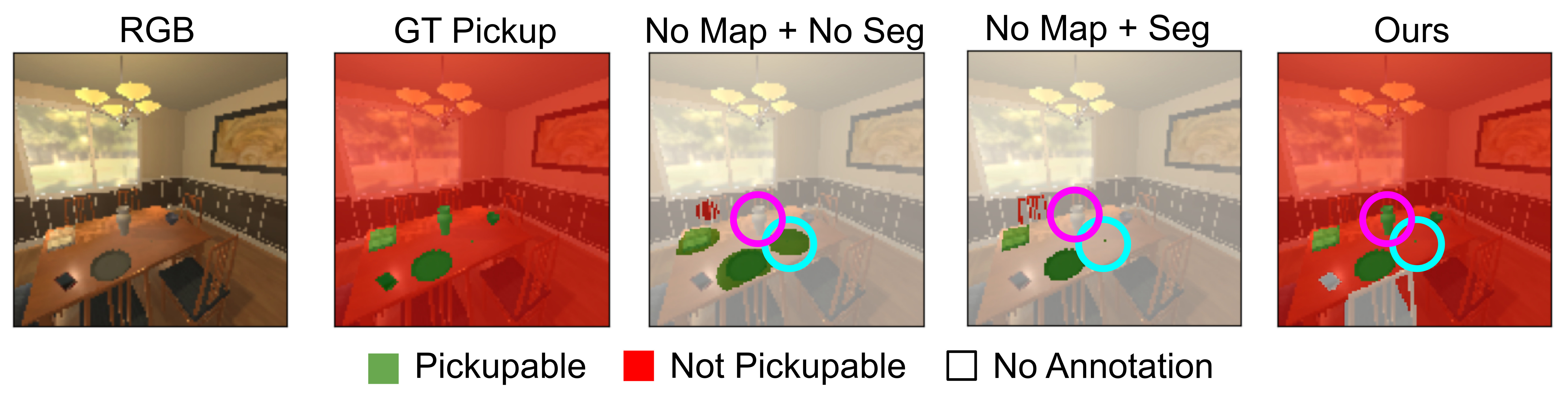}
    \caption{\textit{Pickupable} annotations for an example frame. The cyan circle shows a small pickupable object for which the No Map + No Seg baseline annotates a fixed big area since it does not use the segmentation mask. The magenta circle shows an object that the baseline methods fail to label but our approach is able to annotate by confidence.}
    \label{figure:annotation}
\end{figure}

In each episode, the object positions, object states, initial position of the agent, and initial camera viewpoint are sampled randomly. At every time step, the agent executes an action from its action space (Section~\ref{sec:interactive_exploration_policy}) and obtains feedback from the simulator on whether or not the action was successful. Fig.~\ref{figure:trainingcurves} shows the behavior of the system over the course of the RL training episodes. For our approach the agent not only interacts with more distinct objects per episode, but it also is more successful when trying out interactions with objects (Fig.~\ref{figure:trainingcurves}a and b). This shows that our approach leads to a more targeted exploration which results in more informative labels. Interestingly, annotation by confidence, which is activated after an empirically determined number of \num{400 000} steps, seems to boost the performance of the Interaction Success Rate since more informed interactions can be executed by leveraging the knowledge of the pretrained affordance model. \\
Fig.~\ref{figure:trainingcurves}c and d show that our approach leads to a higher ratio of annotated pixels compared to the approaches that do not accumulate information in a map. Our approach can annotate more pixels as \textit{pickupable} for multiple reasons: Not only does our agent interact with more objects successfully, it also processes these annotations more efficiently by propagating them to multiple frames using the map. Similarly, the high non-interactable annotation rate of our approach is mostly caused by the annotation of large objects like the floor and walls which can be annotated either through interaction or through annotation by confidence and then re-identified in every frame thanks to the map. Overall this suggests that retaining knowledge about object instances, which is only possible when using a map, provides an advantage over just annotating images by backprojecting interaction points as done by \citep{nagarajan2020learning}. \\
Finally, we evaluate the quality of the resulting affordance predictions over time by looking at the pixel-level score Affordance IoU and the Object-level Accuracy (Fig.~\ref{figure:trainingcurves}e and f). Our approach shows a faster and larger increase of the Affordance IoU than for the baselines, which means that the model quickly learns to predict the affordances accurately. The Object-level Accuracy converges quickly to a high level in all approaches, which is partially due to the fact that the predictions on an object level are more forgiving to pixel misclassifications. \\
To visualize the annotation quality, Figure~\ref{figure:annotation} shows a qualitative comparison between our approach and the ablation baselines. As highlighted by the cyan colored circle, the labeling of the No Map + No Seg baseline especially leads to wrong annotations for very small objects since the labeling sphere has a fixed size. The magenta colored circle highlights an object that is annotated in our approach but not in the baselines. This is due to the annotation by confidence process which leverages the pretrained affordance model to label new objects.

\subsection{Affordance Prediction}
\begin{table}[bt]
\centering
    \caption{Affordance prediction (Median and [10th, 90th] percentile), averaged over 611 frames of the test scenes.}
    \label{tab:affordance_results}
    \scriptsize %
    \begin{tabularx}{\textwidth}{lYYYY}
        \toprule
        & IntExp~\citep{nagarajan2020learning} & No Map + No Seg & No Map + Seg & Ours \\
            \midrule
               \multicolumn{5}{c}{Pick up} \\
        \midrule

    {Precision} & 0.03 [0.00, 0.22]   & 0.00 [0.00, 0.06]  & 0.10 [0.00, 0.75]  &  \textbf{0.20 [0.00, 0.99]}  \\ 
     {Recall} & 0.07 [0.00, 0.51]   &  0.0 [0.00, 0.15]  & 0.22 [0.00, 0.96] &  \textbf{0.37 [0.00, 0.99]}   \\ 
      {F1}  & 0.04 [0.00, 0.26] & 0.00 [0.00, 0.08] & 0.13 [0.0, 0.65] & \textbf{0.22 [0.00, 0.85]}  \\ 
     {OAcc}  & 0.86 [0.74, 1.00]  & 0.82 [0.69, 1.00] & 0.87 [0.77, 1.00] &  \textbf{0.89 [0.78, 1.00]}  \\ 
        \midrule
        
     \multicolumn{5}{c}{Push} \\
     \midrule
      {Precision} & 0.36 [0.06, 0.68]  &  0.30 [0.09, 0.52]& 0.52 [0.15, 0.85] & \textbf{0.99 [0.23, 1.00]}  \\
      {Recall} & 0.52 [0.24, 0.69] &  \textbf{0.93 [0.52, 1.0]} & 0.83 [0.12, 0.99] & 0.48 [0.01, 0.96]  \\
       {F1} & 0.41 [0.09, 0.63]  &  0.44 [0.14, 0.67] & 0.58 [0.12, 0.83]  &   \textbf{0.60 [0.01, 0.96]}  \\
         {OAcc}  & 0.65 [0.47, 0.85]  & 0.70 [0.50, 0.86]&  \textbf{0.80 [0.60, 0.92]} &  0.76 [0.57, 0.92]  \\ 
    \bottomrule
    \end{tabularx}
\end{table}

\begin{figure*}[tb]
    \centering
    \includegraphics[width=.97\textwidth]{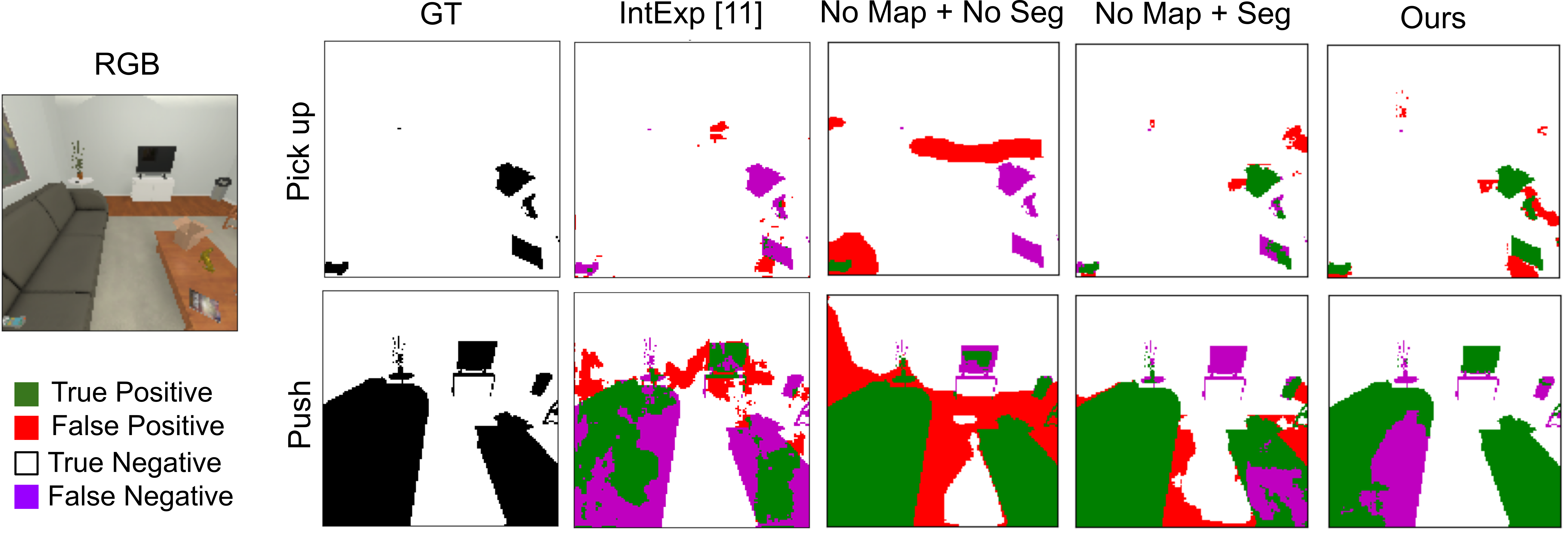}
    \caption{Affordance predictions for an example test frame. For the \textit{pick up} affordance our approach results in more true positives and fewer false negatives. Similarly, for the \textit{push} affordance our approach results in fewer false positives.}
    \label{figure:evaluation_aff_qualitative}
\end{figure*}

\begin{table}[tb]
\centering
\caption{Interaction Success Rate, calculated over the 236 objects of the test scenes.}
    \label{tab:affordance_results_interaction}
    \scriptsize %
    \begin{tabularx}{\textwidth}{lYYYYYYYY}
        \toprule
        & \multicolumn{2}{c}{IntExp~\citep{nagarajan2020learning}} & \multicolumn{2}{c}{No Map + No Seg}  & \multicolumn{2}{c}{No Map + Seg} & \multicolumn{2}{c}{Ours}\\
        \midrule
               &Pick up&Push&Pick up&Push&Pick up&Push&Pick up&Push\\
        \midrule

    {Accuracy }& 0.72 &0.58 & 0.69 &\textbf{0.76} & 0.75 &0.72 &  \textbf{0.81} &0.61 \\ 
     {Precision }& 0.75 &0.76 &  0.50 &0.77 &0.76 &\textbf{0.94} &  \textbf{0.85} &0.90 \\ 
      {Recall }& 0.12 &0.51 & 0.03 &\textbf{ 0.91} & 0.30 &0.60 & \textbf{0.47} &0.43 \\ 
     {F1 }& 0.21 &0.61 & 0.05 &\textbf{0.83} & 0.43 &0.73 &  \textbf{0.60} &0.58 \\
   \bottomrule

    \end{tabularx}
\end{table}

In this section, we evaluate the performance of the learned affordance model in comparison to our ablations as well as to the state-of-the-art work IntExp~\citep{nagarajan2020learning}. 
To evaluate the frame-wise affordance predictions, we created a test dataset by manually steering the agent through five unseen iTHOR living room scenes and recording each frame of the trajectories, resulting in 611 frames. In addition to the previously introduced Object-level Accuracy (OAcc), we use Precision, Recall, and F1 Score. Table~\ref{tab:affordance_results} shows that for the \textit{pick up} affordance our approach outperforms the baseline and the ablations for all metrics. As expected, the IntExp~\citep{nagarajan2020learning} method and the No Map + No Seg ablation behave similarly, since the ablation was designed to mimic the assumptions of~\citep{nagarajan2020learning}. They obtain low scores in precision and recall, which can be explained by the fact that the pickupable objects are usually sparse and small and therefore difficult to learn with the sphere-based annotation strategy. The comparison of the ablations shows clearly that the usage of both the segmentation masks and the map have a positive impact on all metrics. Fig.~\ref{figure:evaluation_aff_qualitative} shows the qualitative prediction results for one example test frame. The pickupable objects on the table are mostly correctly identified by our approach, whereas the IntExp~\citep{nagarajan2020learning} and the No Map + No Seg approaches misclassify all of these small pickupable objects. While our approach predicts some false positives on the table surface, those will tend to be averaged out when evaluating the predictions on an object-level.
For the \textit{push} affordance, Table~\ref{tab:affordance_results} shows a trade-off between precision and recall. Our approach reaches a precision of 0.99, which means that it has a very low number of false positives. The baseline and the ablations on the other hand have a lower precision but a higher recall instead, meaning they are overly likely to predict \textit{pushable}. This is clearly visible in Fig.~\ref{figure:evaluation_aff_qualitative}, where the baseline approaches wrongly predict parts of the floor as pushable, while our approach only labels pixels that belong to pushable objects. Interestingly, in terms of object-level accuracy the No Map + Seg ablations outperforms our approach, which could be caused by the fact that for larger objects the misclassification of a few pixels is averaged out on an object-level.\\
Finally, we also evaluate the Interaction Success Rate to show whether the trained affordance model can correctly predict the feasibility of an interaction. We collected a dataset by placing the robot in front of every object in the test scenes and attempting the \textit{pick up} and \textit{push} actions, storing the outcome as ground truth labels. The affordance prediction modules are queried on the image frame and an object level prediction is calculated as the mean of the pixel-wise predictions of the object. In Table~\ref{tab:affordance_results_interaction} we report the Accuracy, Precision, Recall and F1 score of the predicted object-wise interaction labels compared to the ground truth labels. The results show that for the \textit{pick up} affordance our approach outperforms the baselines. For the \textit{push} affordance, the No Map + No Seg baseline achieves a higher Recall and Accuracy than our approach, while the Precision is lower. Similar to the results from Table~\ref{tab:affordance_results}, this indicates that they are overly likely to predict \textit{pushable}, which could be caused by the spherical annotations spilling over onto other objects.

\section{Conclusion and Future work}
\label{sec:conclusion}

In this work, we proposed a reinforcement learning pipeline to learn robot-centric affordances from interactions. Our approach allows an interactive exploration agent to learn how to efficiently explore an unseen environment while collecting a robot-specific dataset to train an object affordance model. To enable the agent to identify object instances and track object movements, we integrated an object-level map into our pipeline. This allows the agent to re-identify different object instances, thereby obtaining denser annotations from different viewpoints. Additionally, leveraging the information from the object-level map during exploration increases the interaction success rate and efficiency, leading to higher-quality training data. As a result, our method successfully learns to predict robot-centric object affordances.

The next step would be to transfer the learned affordance model to a real-world scenario. Given the domain gap of both the sensor data and the robot morphology, we envision a fine-tuning step in the real-world. We propose to use the pretrained affordance model and RL policy from simulation to collect additional interaction data in the real-world, using a modular robot-specific skill library to execute the high-level interaction commands. Another future step could be to introduce additional actions such as \textit{break} or \textit{open}. While our architecture in general is modular and easily allows adding additional affordances, our chosen TSDF++ representation would have to be developed further to also handle articulated, separable or deformable objects.\\

\AtNextBibliography{\footnotesize} 
\printbibliography

\end{document}